\documentclass[letterpaper, 10 pt, conference]{ieeeconf}  

\IEEEoverridecommandlockouts                              

\usepackage{times}
\usepackage{color, colortbl, xcolor}

\makeatletter
\let\NAT@parse\undefined
\makeatother

\usepackage{multicol}
\usepackage{mathtools}
\usepackage[bookmarks=true]{hyperref}
\usepackage{amsfonts}
\usepackage{amsmath}
\usepackage{subcaption}
\usepackage{multicol}
\usepackage[pdftex]{graphicx}   
\usepackage[ruled,vlined,linesnumbered]{algorithm2e}
\usepackage[noend]{algpseudocode}
\usepackage{cancel}

\usepackage[textfont=md,font=footnotesize]{caption}

\newcommand{\dfknote}[1]%
    {\textcolor{blue}{[DFK: #1]}}
\newcommand{\shnote}[1]%
    {\textcolor{orange}{[SH: #1]}}
\newcommand{\vrnote}[1]%
    {\textcolor{purple}{[VR: #1]}}    
\newcommand{\remove}[1]%
    {\textcolor{red}{#1}}

\newcommand{\example}[1]%
{
  \textbf{Running example:}
  \textit{#1}
}

\newtheorem{proposition}{Proposition}

\newcommand{\roll}{\phi}
\newcommand{\pitch}{\theta}
\newcommand{\yaw}{\psi}
\newcommand{\thrust}{T}

\newcommand{\R}{\mathbb{R}}
\newcommand{\cset}{\mathcal{U}}

\newcommand{\dset}{\mathcal{D}}

\newcommand{\rset}{\mathcal{R}}
\newcommand{\tvar}{t}
\newcommand{\tstate}{s} 
\newcommand{\rstate}{r} 
\newcommand{\rtraj}{\xi}

\newcommand{\mc}{\mathcal}

\usepackage{balance}

\usepackage[%
    style=numeric-comp,
    sorting=none,
    backend=biber,
    sortcites=true,
    doi=false,
    firstinits=true,
    hyperref,
    isbn=false,
    eprint=false,
    maxcitenames=3, 
    minbibnames=3, 
    maxbibnames=4, 
    block=none]
    {biblatex}
    
    \renewbibmacro{in:}{}
    \AtEveryBibitem{%
  	\clearlist{language}%
  	\clearfield{pages}%
	}
\bibliography{references}

\title{\LARGE \bf
A Classification-based Approach for Approximate Reachability
}

\author{Vicen\c{c} Rubies-Royo, David Fridovich-Keil, Sylvia Herbert and Claire J. Tomlin
\thanks{*This research is supported by an NSF CAREER award, the Air Force Office of Scientific Research (AFOSR), NSF's CPS FORCES and VeHICaL projects, the UC-Philippine-California Advanced Research Institute, the ONR MURI Embedded Humans, a DARPA Assured Autonomy grant, and the SRC CONIX Center.}
\thanks{The authors are with the Electrical Engineering and Computer Sciences Dept.,
University of California, Berkeley, USA.
{\tt\small\{vrubies, dfk, sylvia.herbert, tomlin\}@berkeley.edu}}
}

\begin{document}

\maketitle
\thispagestyle{empty}
\pagestyle{empty}

\begin{abstract}
Hamilton-Jacobi (HJ) reachability analysis has been developed over the past decades into a widely-applicable tool for determining goal satisfaction and safety verification in nonlinear systems. While HJ reachability can be formulated very generally, computational complexity can be a serious impediment for many systems of practical interest. Much prior work has been devoted to computing approximate solutions to large reachability problems, yet many of these methods may only apply to very restrictive problem classes, do not generate controllers, and/or can be extremely conservative. In this paper, we present a new method for approximating the optimal controller of the HJ reachability problem for control-affine systems. While also a specific problem class, many dynamical systems of interest are, or can be well approximated, by control-affine models. We explicitly avoid storing a representation of the reachability value function, and instead \emph{learn} a controller as a sequence of simple binary classifiers. We compare our approach to existing grid-based methodologies in HJ reachability and demonstrate its utility on several examples, including a physical quadrotor navigation task. 


\end{abstract}

\section{Introduction}
\label{sec:introduction}

Hamilton-Jacobi (HJ) reachability analysis has proven to be a powerful tool for offline safety verification of nonlinear systems \cite{lygeros1999hybrid, Mitchell2005}. 
The result of such analysis is typically a set of states from which a dynamical system can satisfy a property of interest, and a corresponding controller. These 
could be used, for example, to guarantee that an aircraft will always remain at the proper altitude, heading, and velocity despite uncertain wind conditions. 
While extensive prior work has developed both the theory of reachability analysis and practical tools to compute these sets and controllers \cite{bansal2017hamilton}, numerical approaches to HJ reachability suffer from the ``curse of dimensionality.'' That is, they are unable to cope with ``high'' dimensional system dynamics without large sacrifices in accuracy.  Unfortunately, here 
``high'' means more than five dimensions, which effectively precludes these tools from being used in many key robotics and control applications.

In this paper, we present an approximate dynamic programming approach to mitigating the curse of dimensionality in HJ reachability for control-affine systems. The core idea of our method is to exploit the structure of control-affine systems to avoid computing and storing the large tabular value function used in traditional HJ reachability. For the systems considered here, the control problem at each time step reduces to a tractable set of \textit{binary} classification problems. Importantly, the number of binary classifiers required at each time step is independent of the state space dimension.

Our method yields conservative goal satisfaction and safety guarantees for (a) systems with only control and no disturbance, and (b) systems where we can obtain the worst-case disturbance policy independently, e.g. analytically. 
\begin{figure}
\centering
\includegraphics[width=0.9\linewidth]{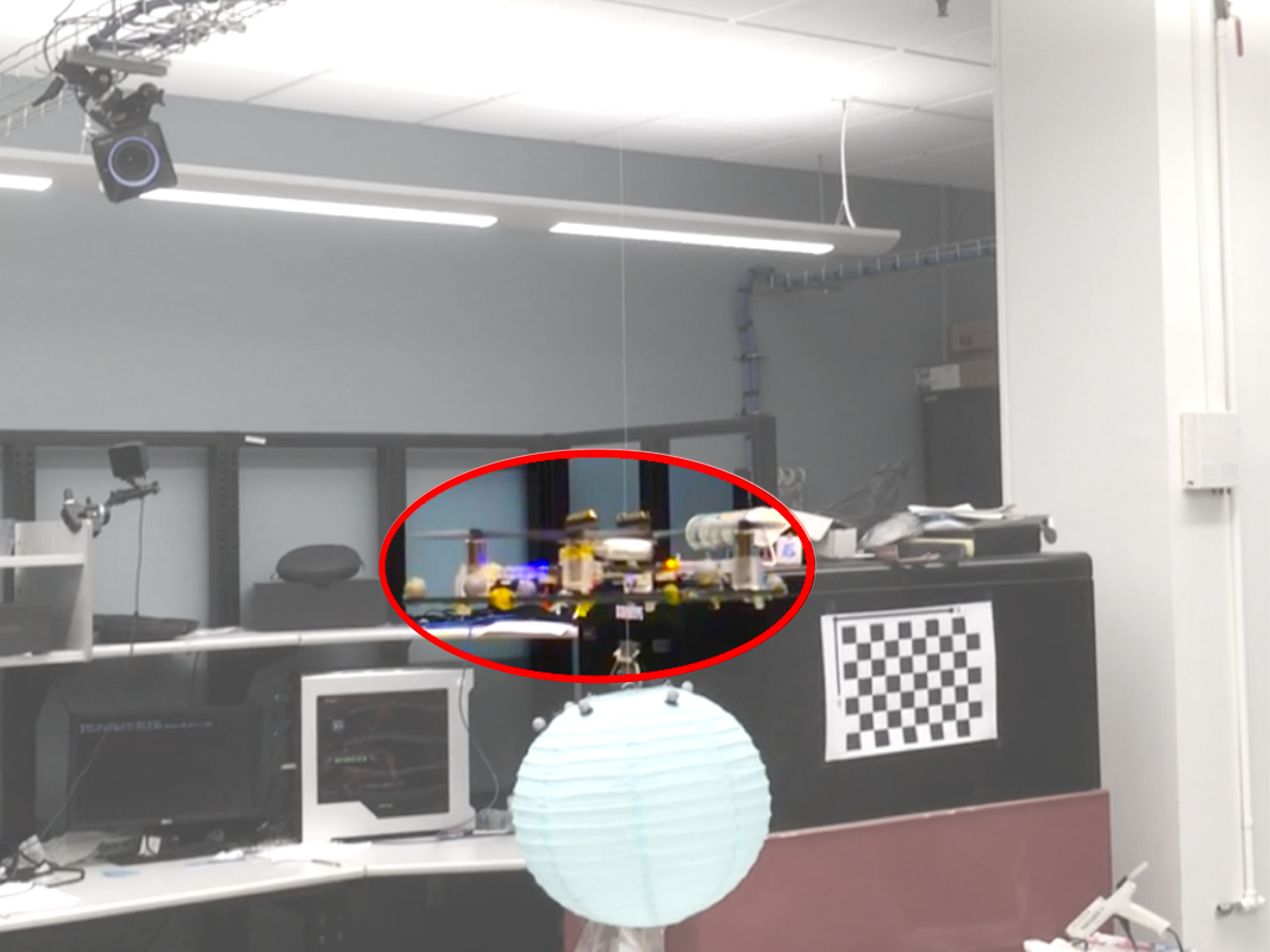}
\caption{Hardware demo of a quadrotor navigating around obstacles (blue lantern) in a motion capture arena using our classification-based controller.
}
	\label{fig:slice_overlay}
	\vspace{-0.65 cm}
\end{figure}
We validate our approximate reachability approach against current grid-based tools using two simulated scenarios, and also test it on a real-time hardware test bed, using a Crazyflie 2.0 quadrotor in a motion capture room shown in Fig.~\ref{fig:slice_overlay}.



\section{Background}
\label{sec:background}

\subsection{Reachability Analysis}
Hamilton-Jacobi (HJ) reachability analysis solves an important class of optimal control problems and differential games. These tools are typically used \textit{offline} to perform theoretical safety analysis and provide goal satisfaction guarantees for nonlinear systems. 
Applications include collision avoidance \cite{Mitchell2005,chen2016robust}, vehicle platooning \cite{Chen15b}, administering anesthesia \cite{kaynama12}, and others \cite{Bayen07,Huang11, Ding08}. 
We can characterize any reachability method (including HJ reachability) according to the following criteria: (a) generality of system dynamics, (b) computation of control and/or disturbance policies, (c) flexibility in representation of sets, and (d) computational scalability. Traditional grid-based HJ reachability methods perform well for the first three criteria, but suffer from poor computational scalability. Recent work has investigated decomposing high-dimensional systems for reachability \cite{Chen2016DecouplingExact,Chen2016DecouplingApprox}; nevertheless, grid-based HJ reachability is often
intractable for analyzing coupled high-dimensional and/or multi-agent systems.

Other reachability methods are more scalable but require linear or affine system dynamics. Such methods may require representing sets using approximative shapes (e.g. polytopes, hyperplanes) \cite{frehse11, greenstreet98, kurzhanski00, kurzhanski02, maidens13}, or not account for control and disturbance inputs \cite{nilsson16}.  
More complex dynamics can be handled by the methods in \cite{althoff15, chen13, dreossi16, frehse11,majumdar14}, but may be less scalable or unable to represent complex set geometries. 

Traditional HJ reachability methods represent the value function directly over a grid, which implicitly specifies the reachable (or avoid, reach-avoid) set, the optimal controller, and if needed, the optimal disturbance. By contrast, in this work, we will compute an approximation of the optimal controller and disturbance directly. Equipped with these approximations, we can compute estimates of the value function and the reachable sets by simulating the known system dynamics with the learned control and disturbance policies.
If a set representation is also required (e.g. for visualization), a grid may be populated using simulated data.

\subsection{Neural Networks Applied to Control Systems}

Feedforward neural networks are a type of parametric function approximator constructed as a composition of nonlinear functions. 
Recently, neural networks have become popular for high-dimensional control tasks. In deep reinforcement learning, for example, neural networks have been employed to learn controllers for complex robotic manipulation tasks, e.g. unscrewing a bottle cap and inserting a peg in a slot \cite{levine2016end,kahn2017plato,nagabandi2017neural,florensa2017reverse}. The control theory literature also includes examples in which neural networks have been successfully employed to find approximate solutions optimal control problems or to learn dynamical system models \cite{bansal2016learning,chen2008generalized,potovcnik2008model,zhang2009neural,dai2014dynamic}.

Neural networks have also been used for approximate reachability analysis \cite{jiang2016using,djeridane2006neural}. Though conceptually related to these approaches, our method differs in that it exploits the structure of control-affine systems to cast the optimal control problem into a (repeated) classification problem. These neural net classifiers can then be used, under some conditions, for verification---i.e. they can be used to provide safety and/or goal satisfaction \textit{guarantees}. 

\section{HJ Reachability Problem Formulation}
\label{sec:forumlation}

Consider a differential game between two players described by the time-invariant system with state $\tstate\in \mc{S} \subset \R^n$ evolving according to the ordinary differential equation:

\begin{equation}
\label{eq:dyn}
\begin{aligned}
\dot{\tstate} &= f(\tstate, u, d), \quad \tvar \in [-T, 0], \\
\end{aligned}
\end{equation}

\noindent where $u \in \mathcal{U} \subset \mathbb{R}^{N_u}$ is the control input and $d \in \mathcal{D} \subset \mathbb{R}^{N_d}$ is the disturbance, which could be due to wind, an adversarial player, etc. Note that we start at an initial \textit{negative} time $-T$ and move towards a final time of $0$. This is because in HJ reachability we typically propagate a value function (defined below) backward in time.
We assume the dynamics $f$
are uniformly continuous, bounded, and Lipschitz continuous in $\tstate$ for fixed $u(\cdot),d(\cdot)$. We define trajectories of this system as $\rtraj(t;\tstate,-T,u(\cdot),d(\cdot))$. The input to this trajectory function is the current time $t$, and it is parameterized by the initial state $\tstate$, the initial time $-T$, and given control and disturbance signals. The output is the state at time $t$.
While prior work in HJ reachability \cite{Varaiya67, Evans84, Mitchell2005} assumes that the disturbance at the current time gets the advantage of seeing the controller's action at that same time (i.e. there exists a causal mapping ${\delta: u(\cdot) \rightarrow d(\cdot)}$), as we will see in Sec. \ref{subsec:dyn_prog_classifiers}, this assumption becomes unnecessary in our framework. 

We represent a target set $\mathcal{L}$  that we want to reach as the zero-sublevel set of an implicit surface function $l(\tstate,t)$, which is generally a signed distance function (i.e. $(s,t) \in \mathcal{L} \iff l(s,t) \leq 0$). This can intuitively be thought of as a cost function representing distance to the target. Likewise, we represent a constraint set $\mathcal{G}$ as the zero-sublevel set of a similar implicit surface function $g(\tstate,t)$. Constraint satisfaction is implied by $g(\tstate,t) \leq 0$. 
As in \cite{Fisac15}, the control wants to minimize, and the disturbance wants to maximize, the cost functional:

\begin{equation} \label{eq:value_1}
    \begin{aligned}
    \mathcal{V}(t,\tstate,u(\cdot),d(\cdot)):=\min_{\tau\in[t,0]} \max\{&\\
    l(\rtraj(\tau;\tstate&,t,u(\cdot),d(\cdot)),\tau),\\
    \max_{q\in[t,\tau]}g(\rtraj(q;\tstate&,t,u(\cdot),d(\cdot)),q)\}.
    \end{aligned}
\end{equation}

\noindent Without the second term in the first \textit{max}, this functional may be interpreted as the minimum distance to the target set ever achieved. If the cost is negative, the set $\mathcal{L}$ was reached within $[t,0]$; otherwise, it was not. The second term then, ensures that any violation of the constraints would override that negative cost. The value of the game is thus given by:

\begin{equation} \label{eq:value_2}
    V(t,\tstate) := \adjustlimits\sup_{\delta[u(\cdot)](\cdot)} \inf_{u(\cdot)}
    \{\mathcal{V}(t,\tstate,u(\cdot),\delta[u(\cdot)](\cdot)) \}.
\end{equation}
This value function characterizes the reach-avoid set, i.e. the set of states from which the controller can drive the system to the target set $\mathcal{L}$ while staying within constraint set $\mathcal{G}$, despite the worst-case disturbance: ${\mathcal{RA}_t := \{\tstate : V(t,\tstate)\leq 0\}}$. 

Finally, at a given time $t$, it is known \cite{liberzon2011calculus} that the optimal control $u^*$ and disturbance $d^*$ must satisfy
\begin{align}
\label{eqn:maximum_principle}
    u^*, d^* = \arg\adjustlimits\inf_{u' \in \cset}\sup_{d' \in \dset} \nabla_s V(t, s)^T f(s, u', d') .
\end{align}

 Approaches in \cite{Mitchell2005,Fisac15,Bokanowski10, barron90} are compatible with well-established numerical methods \cite{LSToolbox, Mitchell08, Osher02, sethian96}. However, these approaches compute $V(t, \tstate)$ on a grid, and they quickly become intractable as the dimensionality of the problem increases. 





\section{Classifier-based Approximate Reachability}
\label{sec:NN}

In this section we introduce our classification-based method for approximating the optimal control of HJ reachability when the dynamics are control-affine. Even though we will use feedforward neural networks to build the classifiers, it is possible to use other methods (e.g. SVM, decision trees). Ultimately, the choice of the classifier determines how conservative the results of the procedure will be. We leave a full investigation of classifier performance for future work.

\subsection{Control/Disturbance-Affine Systems} \label{sec:control-affine}
A control/disturbance-affine system is a special case of (\ref{eq:dyn}) of the form

\begin{equation}
\label{eq:aff_dyn}
\begin{aligned}
\dot{\tstate} &= \alpha(\tstate) + \sum_{i=1}^{N_u}\beta_i(\tstate)u_i + \sum_{j=1}^{N_d}\gamma_j(\tstate)d_j~,\\
\end{aligned}
\end{equation}
\noindent where $\alpha,\beta_i,\gamma_j: \mathbb{R}^n \to \mathbb{R}^n$. We will assume that both control and disturbance are bounded by interval constraints along each dimension, i.e. $u_i \in [u_{min}^i, u_{max}^i]$ for $i=1,\dots,N_u,$ and $d_j \in [d_{min}^j, d_{max}^j]$ for $j=1,\dots,N_d$.
Observe that when dynamics $f$ are of the form \eqref{eq:aff_dyn}, the objective in \eqref{eqn:maximum_principle} is affine in the instantaneous control $u'$ and disturbance $d'$ at every time $t$. The optimal solution, therefore, lies at one of the $2^{N_u}$ (or $2^{N_d}$) corners of the hyperbox containing $u$ (or $d$). That is, the optimal control and disturbance policies are ``bang-bang''\footnote{For many physical systems, it is preferable to apply a smooth control signal. We note that the bang-bang control resulting from \eqref{eqn:maximum_principle} need only be applied at the boundary of the reach-avoid set.} (we refer the reader to chapter 4 of \cite{liberzon2011calculus}). Furthermore, the optimal values for any $u_i$ or $d_j$ at a certain state and time are mutually independent; therefore, for control/disturbance-affine systems, we can frame the HJ reachability problem \eqref{eq:value_2} as a series of $N_u + N_d$ \textit{binary} classification problems at each time.

\subsection{Dynamic Programming with Binary Classifiers}
\label{subsec:dyn_prog_classifiers}


Algorithm~\ref{alg:PSC} describes the process of learning these classifiers in detail. We begin by discretizing the time-horizon $T$ into small (evenly spaced) intervals of size $\Delta t>0$ in line \ref{alg:disc_t}, and proceed to use the dynamic programming principle backwards in time to build a sequence of approximately optimal control and disturbance policies. In total, the number of classifiers will be $\frac{T}{\Delta t}(N_u + N_d)$. 

\begin{algorithm}[ht!]
 \textbf{Input:} $\dot{\tstate} = f(\tstate, u, d), \mathcal{S}, \mathcal{U}, \mathcal{D}, T, \Delta t, C, N, Trn(\cdot,\cdot)$ \\
 \textbf{Initialize} $\Pi_{u}, \Pi_{d} \gets \{\}$ \\
 \textbf{For} $k=0, \dots, \left\lfloor{T/\Delta t}\right\rfloor$ \\ \label{alg:disc_t}
  \Indp
     \textbf{Initialize} $P, U^*, D^* \gets \{\}$  // No training data.\\
    \textbf{For} $q = 1, \dots, N$\\ \label{alg:N_states}
    \Indp
        \textbf{Sample} $\tstate \sim \text{Unif}\{\mc{S}\}$ \\ \label{alg:S_uni}
        \textbf{Initialize} $u^*, d^* \gets u_{min},d_{min}$ \\ \label{alg:opt_init}
        $\hat{s} \gets \xi(-k \Delta t; \tstate, -(k+1)\Delta t, u_{min}, d_{min})$\\
        $\hat{c} \gets C(\hat{s}, \Pi_{u}, \Pi_{d})$\\
        \textbf{For} $i = 1, \dots, N_u$ ~ //  Find best control.\\
        \Indp
            $\hat{u} \gets u_{min}$; $\hat{u}^i \gets u^i_{max}$\\
            $\tstate' \gets \xi(-k \Delta t; \tstate, -(k+1)\Delta t, \hat{u}, d_{min})$\\
            \textbf{If} ($C(\tstate', \Pi_{u}, \Pi_{d}) < \hat{c}$): $u^*_{i} \gets u^i_{max}$\\
        \Indm
        \textbf{For} $j = 1, \dots, N_d$ ~ // Find best disturbance.\\
        \Indp
            $\hat{d} \gets d_{min}$; $\hat{d}^i \gets d^i_{max}$\\
            $\tstate' \gets \xi(-k \Delta t; \tstate, -(k+1)\Delta t, u_{min}, \hat{d})$\\
            \textbf{If} ($C(\tstate', \Pi_{u}, \Pi_{d}) > \hat{c}$): $d^*_{i} \gets d^i_{max}$\\ \label{alg:opt_end}
        \Indm
        $U^* \gets \{U^*, u^*\}$ // Record control.\\ \label{alg:fill_data_beg}
        $D^* \gets \{D^*, d^*\}$ // Record disturbance.\\
        $P \gets \{P, \tstate\}$ // Record state.\\ \label{alg:fill_data_end}
    \Indm
    // Train new classifiers. Add them to overall policy. \\
     $\Pi_{-(k+1) \Delta t}^u \gets Trn(P,U^*), \Pi_{u} \gets \{\Pi_{u},\Pi_{-(k+1) \Delta t}^u\}$\\ \label{alg:second_t}
     $\Pi_{-(k+1) \Delta t}^d \gets Trn(P,D^*), \Pi_{d} \gets \{\Pi_{d},\Pi_{-(k+1) \Delta t}^d\}$\\ \label{alg:o_last}
 \Indm 
\textbf{Return} $\Pi_{u},\Pi_{d}$
\caption{Learning policies and disturbances \label{alg:PSC}}
\end{algorithm}

At an intermediate time $t<0$, we will have already obtained the binary classifiers for the control and disturbance policies from $t + \Delta t$ to 0: $\Pi^u_{(t + \Delta t):0}$
and $\Pi^d_{(t + \Delta t):0}$. Here, $\Pi^{u}_\tau$ and $\Pi^{d}_\tau$ each denote a \textit{set of classifiers} for the discrete time step $\tau$ (i.e. $|\Pi^c_{\tau}| = N_u$ and $|\Pi^d_{\tau}| = N_d$). We now define the function $C$, which computes the cost \eqref{eq:value_2} if control and disturbance acted according to these pre-trained policies:
\begin{align}
\label{eqn:cost}
    C(\tstate, \Pi^u_{(t + \Delta t):0}, &\Pi^d_{(t + \Delta t):0}) :=  \mathcal{V}(t,\tstate,u(\cdot),d(\cdot)) ,
\end{align}
\noindent where, due to our discretization, control and disturbance are piecewise constant over time, i.e. $u(t) = \Pi^u_{\tau}$ and $d(t) = \Pi^d_{\tau}$ for $t \in [\tau, \tau + \Delta t)$ and all discrete time steps $\tau$. 

At time $t$, we can determine for some arbitrary state $s$ the optimal control and disturbance as follows. First, compute the cost of applying ${u_{min} = (u_{min}^0, \dots, u_{min}^{N_u})}$ and ${d_{min} = (d_{min}^0, \dots, d_{min}^{N_d})}$ from $t$ to $t + \Delta t$; that is, let $\hat c = C\big(\xi(t+\Delta t; s, t, u_{min}, d_{min}),  \Pi^u_{(t + \Delta t):0}, \Pi^d_{(t + \Delta t):0}\big)$. Now, separately for each component $i$ of $u$ (and likewise for $d$), set $u^i(t) = u_{max}$ and compute the cost. If the cost is less than (resp. greater than, for disturbance) $\hat c$, then this is the optimal control (resp. disturbance) in dimension $i$ at time $t$. This corresponds to lines \ref{alg:opt_init}-\ref{alg:opt_end}.

Equipped with this procedure for computing \textit{approximately optimal}\footnote{Approximately optimal, since we compute policies at time $t$ based on previously trained control and disturbance policies for $\tau > t$.} control and disturbance actions, we record the computed state-action pairs (lines \ref{alg:fill_data_beg}-\ref{alg:fill_data_end}) for $N$ states sampled uniformly over $\mathcal{S}$\footnote{While other distributions could be used, in this work we focus solely on uniform sampling. Different sampling strategies may result in different algorithm performance.} (lines \ref{alg:N_states}-\ref{alg:S_uni}). We then train separate binary classifiers for \textit{each component} of $u$ and $d$, and add them to their current set $\Pi^u_\tau$ or $\Pi^d_\tau$. These are finally appended to the time-indexed control and disturbance policy sets $\Pi_u$ and $\Pi_d$ (lines \ref{alg:second_t}-\ref{alg:o_last}). $Trn(\cdot,\cdot)$ denotes a training procedure given state-action pairs. The Appendix contains further details pertaining to how the classifiers were trained.

Two of the main benefits of performing approximate reachability analysis using binary classifiers rather than grids are memory usage and time complexity. The memory footprint of medium-sized neural networks of the sort used in this paper can be on the order of $10^3$ parameters or $\sim10$ Kb, as opposed to $\sim10$ Gb for dense grids of 4D systems. In our experience, Algorithm \ref{alg:PSC} typically terminates after an hour for the 6D and 7D systems presented in Sec. \ref{sec:fastrack}, whereas grid-based methods are completely intractable for coupled systems of that size.

\subsection{Special Case: Value Function Convergence}
\label{subsec:remarks}

For some instances of problem \eqref{eq:value_1} and \eqref{eq:value_2} the value function $V(t,s)$ converges: $\lim_{t \to -\infty} V(t,s) = V^*(s)$. From \eqref{eqn:maximum_principle}, the corresponding optimal control and disturbance policies also converge. While in this paper we make no claims regarding convergence of the classifiers to the true optimal policies, our empirical results do suggest convergence in practice (see Fig. \ref{fig:learning_curves}).
When this happens, we denote $\Pi_{-T}^u = \Pi_{-\infty}^u$ (resp. $\Pi_{-T}^d = \Pi_{-\infty}^d$), for $T$ large enough. In practice, the horizon can be progressively increased as needed. A benefit of converged policies is that when estimating $V^*(s)$ we only require the last set of binary classifiers $\Pi_{-\infty}^u$ and $\Pi_{-\infty}^d$, allowing us to store only $N_u + N_d$ classifiers.

\subsection{Summary of Guarantees}
\label{subsec:guarantees}

Algorithm \ref{alg:PSC} returns a set of approximately optimal policies for the control and the disturbance for a finite number of time steps. Recalling \eqref{eqn:cost}, in order to obtain an estimate of the value at a certain state $\tstate$ and time $t$, it suffices to simulate an entire trajectory from that state and time using the learned policies. The value $V^{\Pi_u, \Pi_d}(t,\tstate)$ is the cost of the associated trajectory, measured according to \eqref{eq:value_1}. 

A benefit of working with policy approximators rather than value function approximators is that in the case of no disturbance, the value function induced by the learned control policy will always upper-bound the true value. This means that a reach-avoid set computed via Algorithm~\ref{alg:PSC} will be a subset of the true reach-avoid set.
For reachability problems involving a disturbance, if the optimal disturbance policy is known \textit{a priori}, the same guarantee still applies. However, if the optimal disturbance is unknown and must also be learned, no guarantees can be made 
because the learned disturbance policy will not generally be optimal. We formalize this result with the following proposition.
\begin{proposition}
\label{prop:guarantees}
If we assume (a) no disturbance, or (b) access to a worst-case optimal disturbance policy, then the computed reach-avoid set is a subset of the true set.\\ 
\begin{proof}
First assume no disturbance. Due to the use of function approximators, the control policy $\Pi_u$ will be suboptimal relative to the optimal controller $u^*(\cdot)$, meaning it is less effective at minimizing the cost functional \eqref{eq:value_1}. Therefore, $V^{\Pi_u}(t,\tstate) \geq V(t,\tstate)$. Denoting the neural network reach-avoid sets as $\mathcal{RA}^{\Pi_u}_t := \{\tstate : V^{\Pi_u}(t,\tstate)\leq 0\}$, this inequality implies that $\mathcal{RA}^{\Pi_u}_t \subseteq \mathcal{RA}_t$. \end{proof}

Note that this applies to \emph{all} states $s$ and times $t$, not just those that were sampled in Algorithm~\ref{alg:PSC}. When optimizing over both control and disturbance this guarantee does not hold because the disturbance will generally be suboptimal and therefore not worst-case. However, when provided with an optimal disturbance policy at the onset, we recover the case of optimizing over only control.
\end{proposition}
\section{Examples}
\label{sec:examples}

\begin{figure}[t]
     \centering
    \includegraphics[width=.95\linewidth]{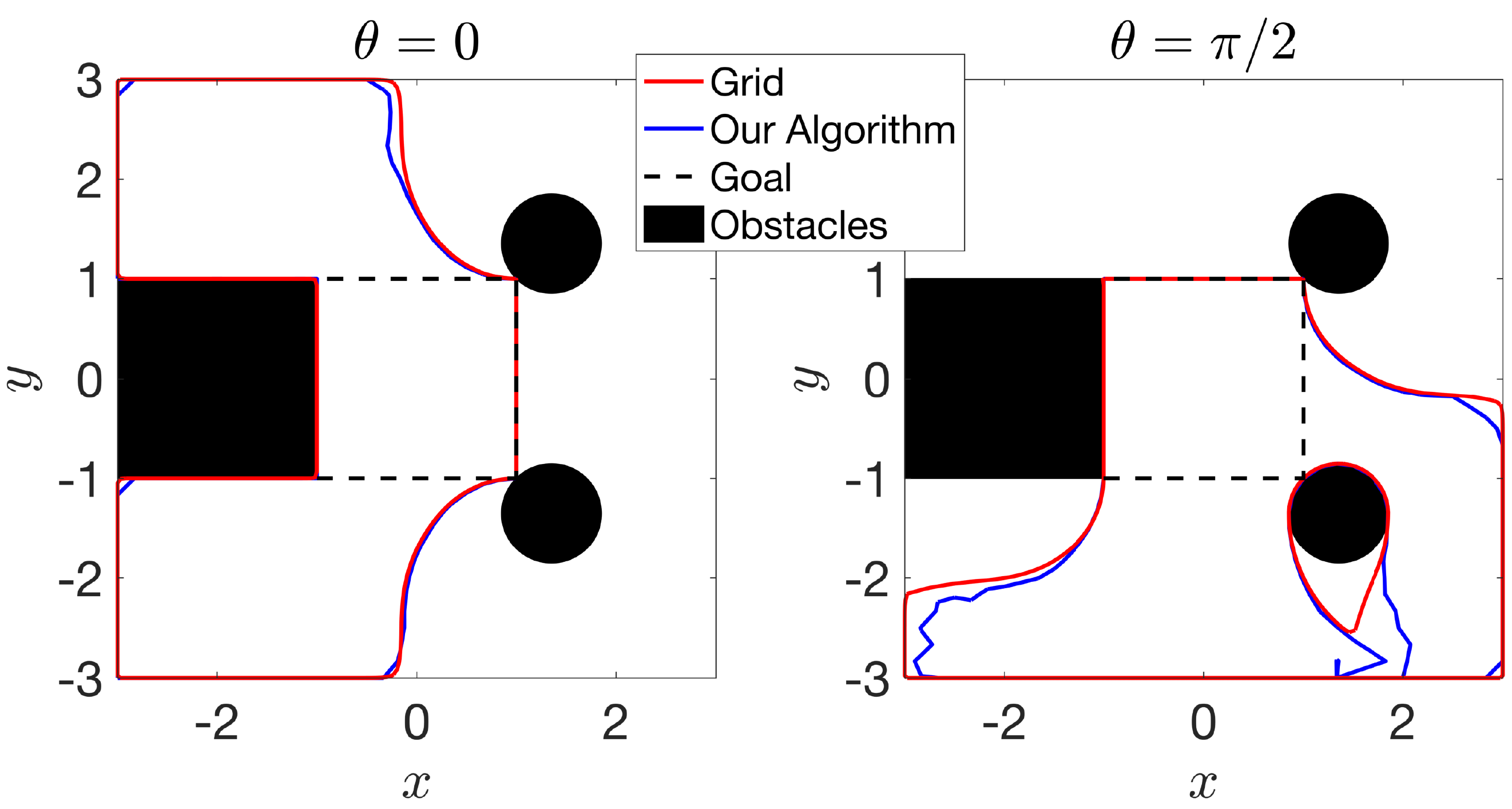}
    \caption{Reach-avoid set computation for the 4D dynamics in \eqref{eqn:4d_unicycle_dynamics} for two different 2D slices and tangential speed $v=1$.}
    \label{fig:4d_unicycle}
    \vspace{-.6cm}
\end{figure}

In this section, we will present two reachability problems \textit{without} disturbances, and compare the results of our proposed method with those obtained from a full grid-based approach \cite{LSToolbox}. In each case, we observe that our method agrees with the ground truth, with a small but expected degree of conservatism. For these examples, the set $\mathcal{L}$ is a box of side-length $2$ centered at $(x,y) = (0,0)$, and $\mathcal{G}$ consists of the outer boundaries (i.e. $\max\{|x|,|y|\}\leq 3$) and the shaded obstacles (Fig.~\ref{fig:2d_point} and Fig.~\ref{fig:4d_unicycle}).

\subsection{2D point}
\label{subsec:2d_mass}

Consider a 2D dynamical system with inputs $u_1 \in [\underline{u}_1, \overline{u}_1]$ and $u_2 \in [\underline{u}_2, \overline{u}_2]$ which evolves as follows: \vspace{-1em}

\begin{equation}
\label{eqn:2d_point_dynamics}
\dot x = u_1, ~
\dot y = u_2
\end{equation}
 
\begin{figure}[t]
    \centering
    \includegraphics[width=.95\linewidth]{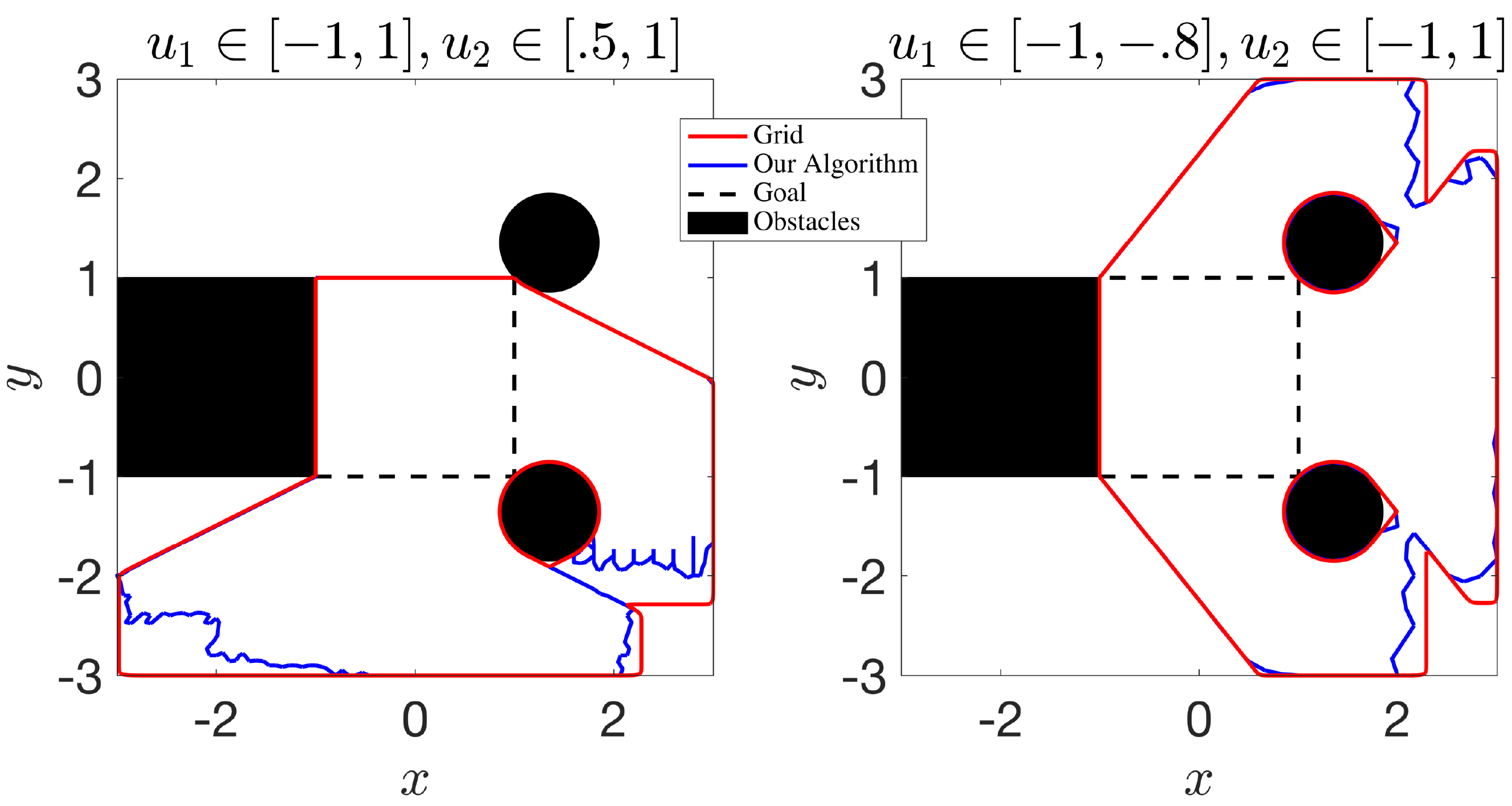}
    \caption{Reach-avoid set computation for 2D dynamics in \eqref{eqn:2d_point_dynamics} for two different control bounds. Sets computed using our method are subsets of the true sets.}
    \label{fig:2d_point}
    \vspace{-.3cm}
\end{figure}

Fig.~\ref{fig:2d_point} shows the reach-avoid sets for two different control bounds. We overlay the sets computed by our method on top of that computed using a dense $121 \times 121$ grid \cite{LSToolbox}. The red set was computed using standard HJ reachability and the blue set was computed using our classification-based method. Points inside the reach-avoid sets represent states from which there exists a control sequence which reaches the target while avoiding all obstacles. As guaranteed in Proposition~\ref{prop:guarantees}, the set computed via Algorithm~\ref{alg:PSC} is always a subset of the ground truth, meaning that every state marked in Fig.~\ref{fig:2d_point} as safe is also safe using the optimal controller. The computation time for the grid-based approach was $20$ seconds, while for the classification-based it was $10$ minutes.

\subsection{4D unicycle}
\label{subsec:4d_unicycle}

Next, we consider a higher-dimensional system representing a 4D unicycle model: 
\begin{align}
\label{eqn:4d_unicycle_dynamics}
\dot s = \begin{bmatrix}
\dot x & \textrm{(x-position)} \\
\dot y & \textrm{(y-position)} \\
\dot \theta & \textrm{(yaw angle)} \\
\dot v & \textrm{(tangential speed)}
\end{bmatrix} = \begin{bmatrix}
v \cos \theta \\
v \sin \theta \\
u_{\omega} \\
u_a
\end{bmatrix}
\end{align}
in which controls are tangential acceleration $u_a \in [0,1]$ and yaw rate $u_{\omega} \in [-1,1]$.
Fig.~\ref{fig:4d_unicycle} shows a computed a reach-avoid set for this system for different 2D slices of the 4D state space on a $121^4$ grid. As expected, our approach yields a conservative subset of the true reach-avoid set. In this case, the computation time for the grid-based approach was $3$ days, while for the classification-based it was $30$ minutes.

\section{Hardware Demonstration: FaSTrack}
\label{sec:fastrack}

In this section we will use our method to compute a controller for a quadrotor. We will be using a trajectory tracking framework (FaSTrack) which is based on a variant of the reachability problem \eqref{eq:value_1}. Unlike Sec. \ref{sec:examples}, we will also be considering a disturbance signal.

\subsection{FaSTrack overview}
\label{subsec:fastrack_overview}
FaSTrack (Fast and Safe Tracking) is a recent method for safe real-time motion planning \cite{herbert2017fastrack}. FaSTrack breaks down an autonomous system into two agents: a simple \textit{planning model} used for real-time motion planning, and a more complicated \textit{tracking model} used to track the generated plan.
To ensure safe tracking, FaSTrack computes the largest \textit{relative} distance between the two models (tracking error), and the planning algorithm uses this result to enlarge obstacles for collision-checking. The computation also provides an optimal feedback controller to ensure that the tracker remains within this bound during planning.

To solve for the largest tracking error in FaSTrack, we set the cost $l(r,t)$ in \eqref{eq:value_1} as the distance to the origin in relative position space. We denote relative states by $\rstate \in \rset \subset \mathbb{R}^{N_r}$ (see Sec.~\ref{subsec:fastrack_precomputation}), and solve a modified form of \eqref{eq:value_1}:
\begin{equation} \label{eq:fastrack_value_1}
    \mathcal{V}(t,\rstate,u(\cdot),d(\cdot)):=\max_{\tau\in[t,0]}
    l(\rtraj(\tau;\rstate,t,u(\cdot),d(\cdot)),\tau)
\end{equation}
Note that there is no constraint function $g(\rstate,t)$. Also, we now take the \emph{maximum} value over time because we want to find the maximum relative distance that could occur between the two models. Finally, observe that in this formulation, the disturbance actually encompasses two separate quantities: the original notion of disturbance (e.g. wind), \emph{and} the planning model's control input, which directly affects the relative state dynamics. Henceforth, policy $\Pi_{-\infty}^d$ will represent the concatenated disturbance and planning algorithm policies.

\begin{figure*}[ht!]
\centering
\begin{subfigure}[t]{0.2\textwidth}
\centering
\includegraphics[width=\textwidth]{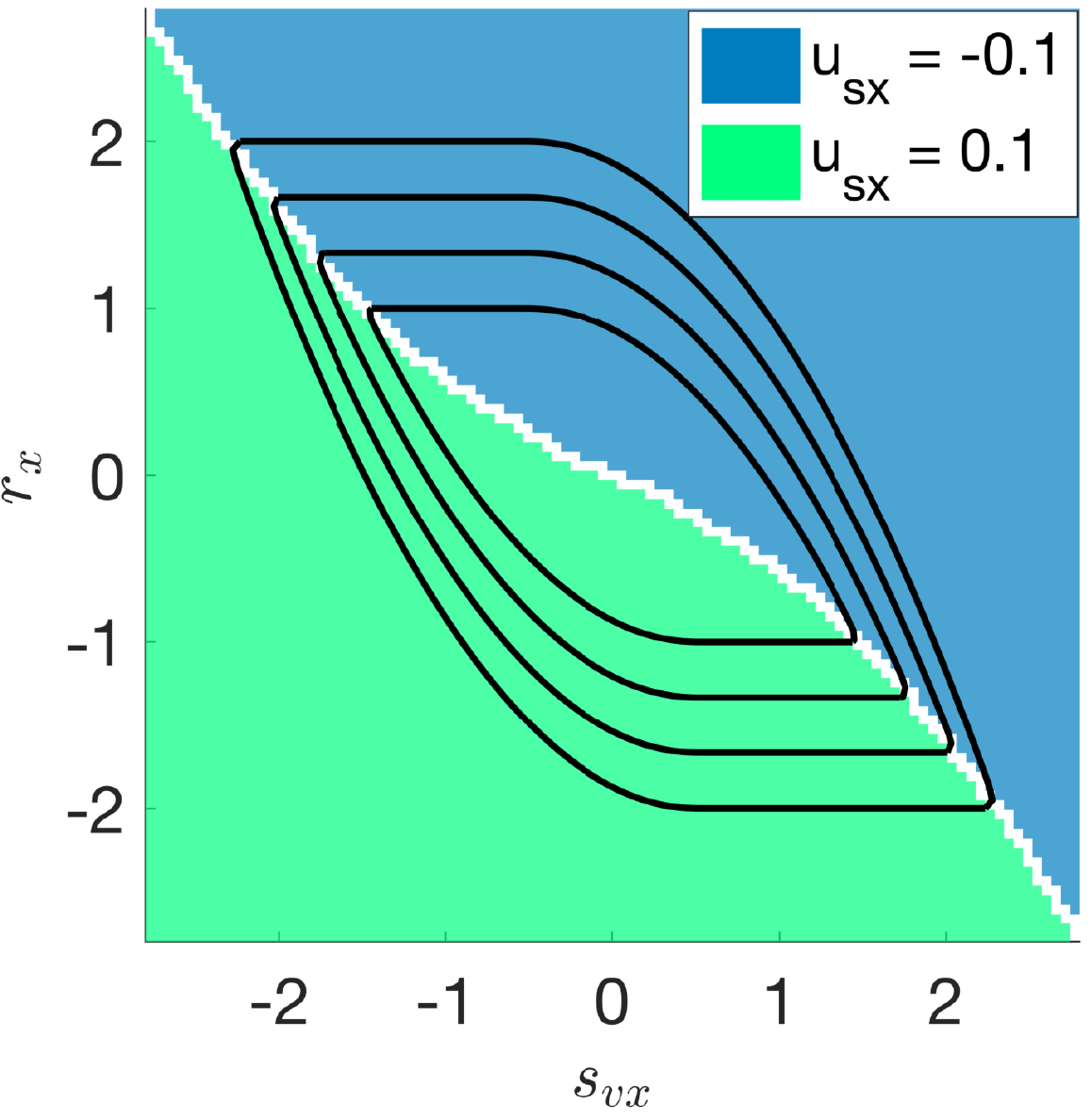}
\caption{Ground truth (grid)}
\label{subfig:grid_x}
\end{subfigure}
\hspace{2em}
~
\begin{subfigure}[t]{0.2\textwidth}
\centering
\includegraphics[width=\textwidth]{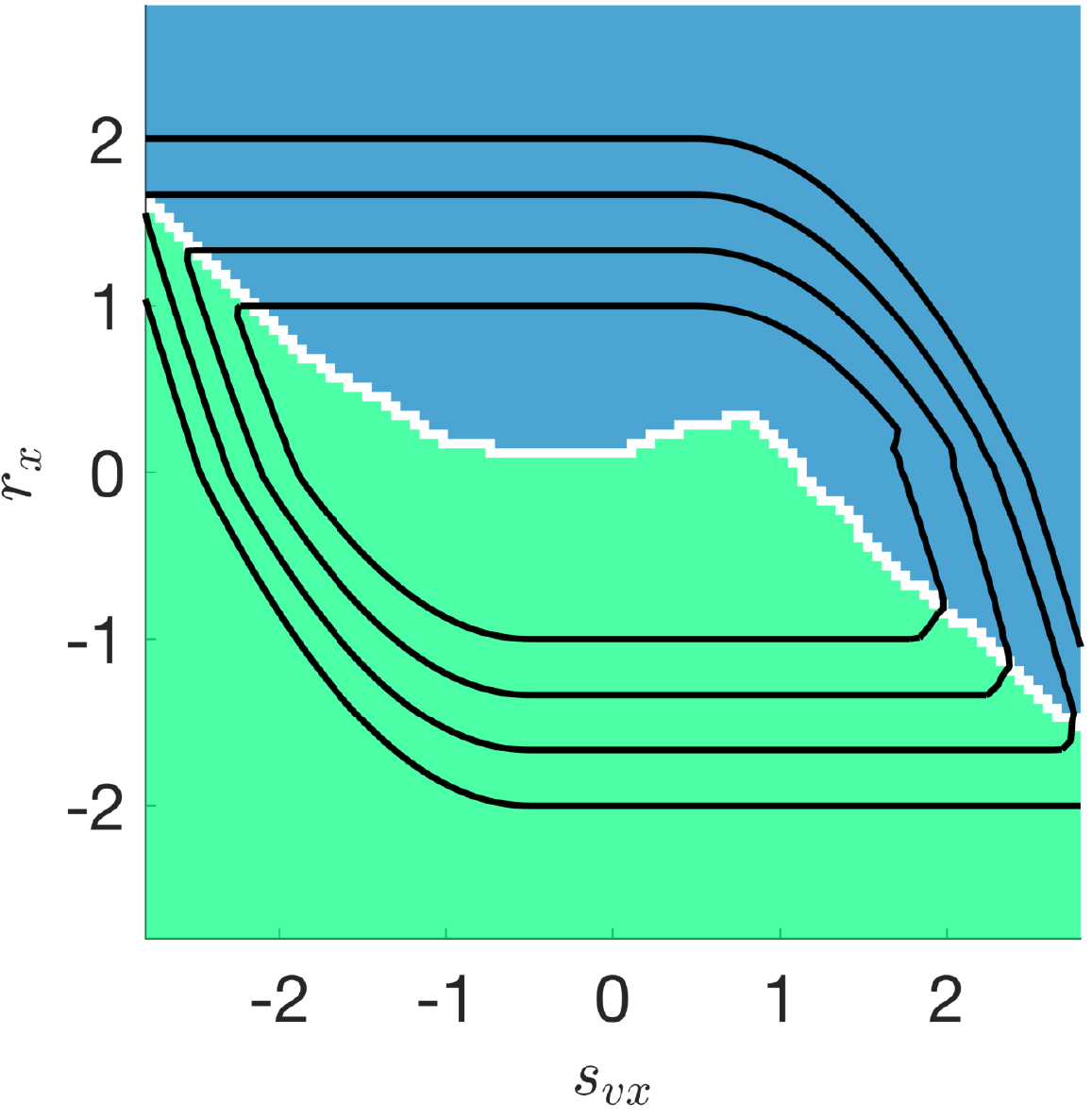}
\caption{Classifier $\Pi_{-\infty}^u$} 
\label{subfig:nn_optdst_x}
\end{subfigure}
\hspace{2em}
~
\begin{subfigure}[t]{0.2\textwidth}
\centering
\includegraphics[width=\textwidth]{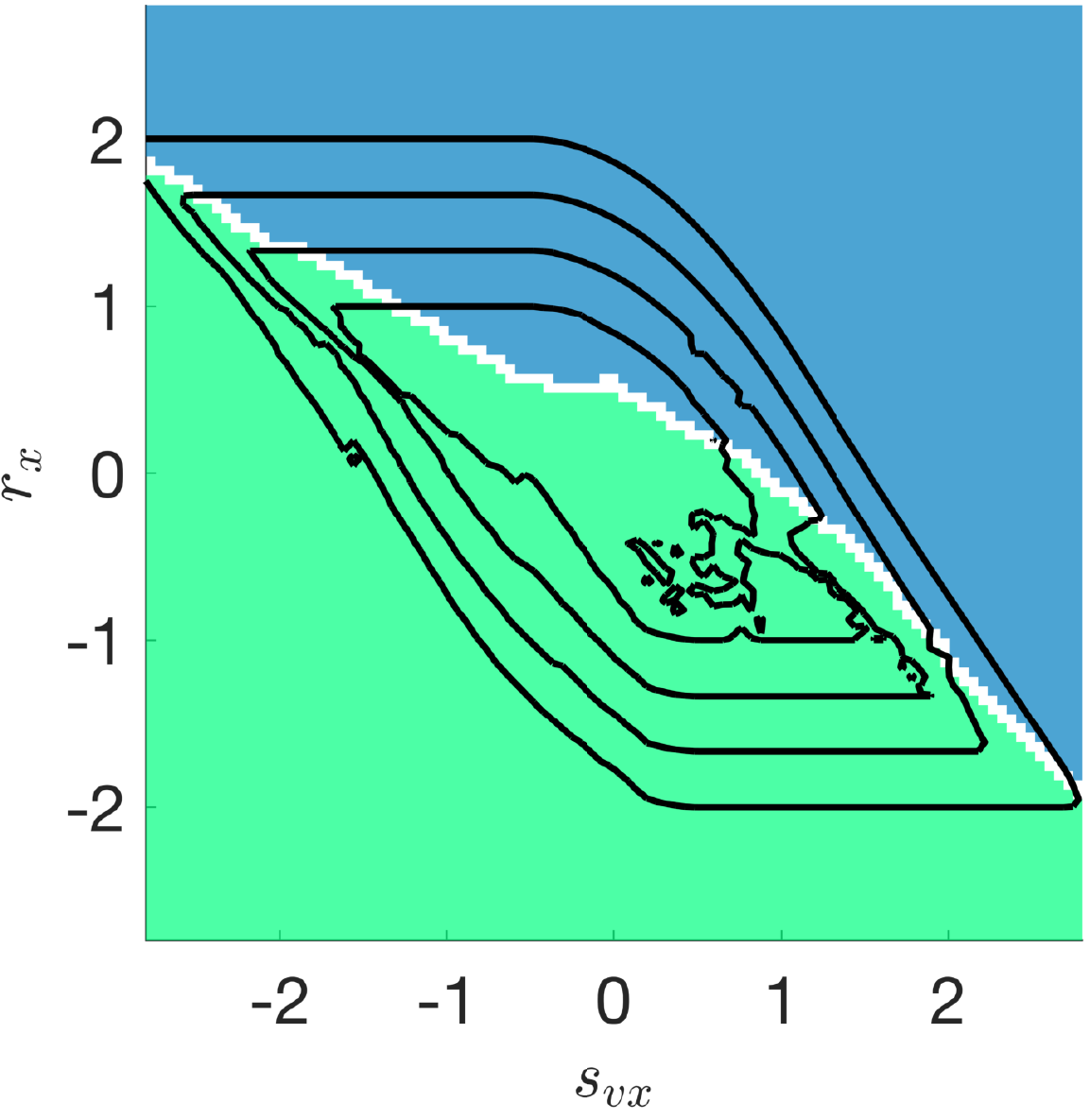}
\caption{Classifiers $\Pi_{-\infty}^u$, $\Pi_{-\infty}^d$}
\label{subfig:nn_lrndst_x}
\end{subfigure}
~
\begin{subfigure}[t]{0.2\textwidth}
\centering
\includegraphics[width=\textwidth]{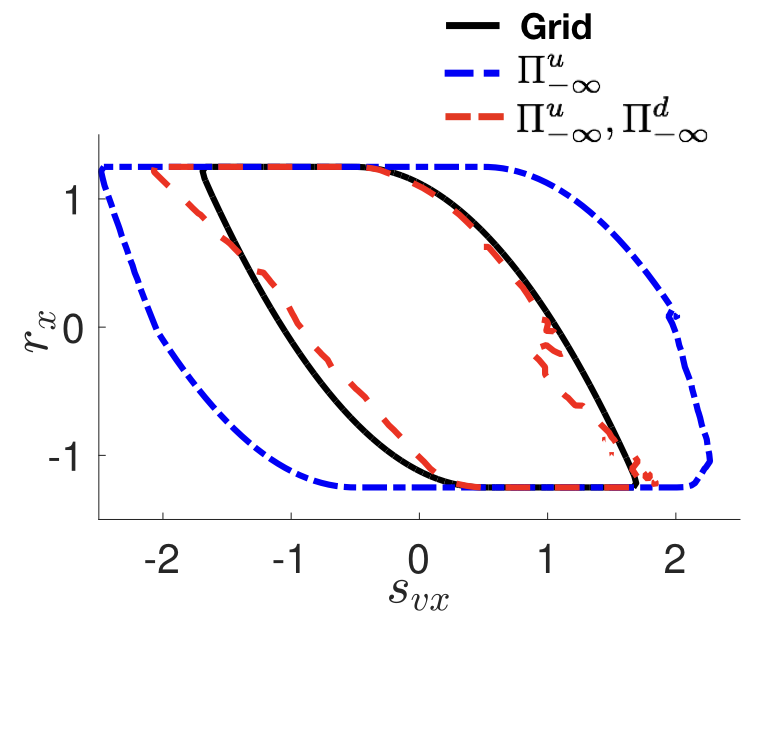}
\caption{Overlaid level curves.}
\label{subfig:fastrack_overlay}
\end{subfigure}
\caption{Level sets of $V^*$ in the $(r_x, v_x)$ states (setting other states to zero) for (a) ground truth grid-based representation, (b) neural network $\Pi_{-\infty}^u$ trained on optimal disturbance policy $d^*(\cdot)$, and (c) neural networks $\Pi_{-\infty}^u$ and $\Pi_{-\infty}^d$ trained jointly. We encode the optimal (learned) control at each state as a different color. (d) Overlay of level sets from (a-c). Our method only yields a conservative result (a superset of the ground truth; see Sec.~\ref{subsec:fastrack_overview}) when $\Pi_{-\infty}^u$ is trained against $d^*(\cdot)$.}
\label{fig:6D_decoupled_x}
\vspace{-0.5cm}
\end{figure*}



Following Section \ref{subsec:guarantees}, when the optimal converged disturbance policy $\Pi_{-\infty}^d$ is known analytically, the policies learned in Algorithm~\ref{alg:PSC} will (by Prop.~\ref{prop:guarantees}) yield a value function which \textit{over-approximates} the optimal value function, i.e. $V^{\Pi_u, d^*}(t,\rstate) \geq V(t,\rstate)$. 
Thus, the maximum relative distance ever achieved between tracking mode and planning model, from any initial relative state, will always be \textit{greater} when using the binary classifier policies than the optimal policy. For safe trajectory tracking, this translates into enlarging obstacles by a larger amount, meaning we still preserve safety.


\subsection{FaSTrack Reachability Precomputation}
\label{subsec:fastrack_precomputation}

We employ Algorithm~\ref{alg:PSC} to find the largest tracking error for two nonlinear models of the \textit{tracking model}, which become control-affine under small angle assumptions. First, we consider a 6D near-hover model which decouples into three 2D subsystems and thus admits a comparison to grid-based methods. Then, we present results for a \textit{fully-coupled} 7D model that cannot be solved exactly using grid-based techniques and use it for quadrotor control. 

\subsubsection{6D Decoupled}
\label{subsubsec:6D_decoupled}

We first consider a 6D quadrotor tracking model and 3D geometric planning model. Here, the quadrotor control consists of pitch ($\pitch$) and roll ($\roll$) angles, and thrust acceleration ($\thrust$), while the planning model's maximum speeds are $b_x, b_y$, and $b_z$ in each dimension. All of our results assume $\roll, \pitch \in [-0.1, 0.1]$ rad, $\thrust - g \in [-2.0, 2.0]~\text{m}/\text{s}^2$, and $b_x = b_y = b_z = 0.25~\text{m}/\text{s}$. We assume a maximum velocity disturbance of $0.25~\text{m}/\text{s}$ in each dimension. The relative position states $(\rstate_x, \rstate_y, \rstate_z)$ and the tracker's velocity states $(\tstate_{vx}, \tstate_{vy}, \tstate_{vz})$ adhere to the following \emph{relative dynamics}:
\begin{align}
\label{eq:Quad6D_decoupled_relative}
\begin{bmatrix}
\dot \rstate_x \\
\dot \rstate_y \\
\dot \rstate_z
\end{bmatrix} = 
\begin{bmatrix}
\tstate_{vx} - d_{vx} - b_x\\
\tstate_{vy} - d_{vy} - b_y\\
\tstate_{vz} - d_{vz} - b_z
\end{bmatrix}~,
\begin{bmatrix}
\dot \tstate_{vx} \\
\dot \tstate_{vy} \\
\dot \tstate_{vz} 
\end{bmatrix} = 
\begin{bmatrix}
g \tan\pitch\\
-g \tan\roll\\
\thrust - g
\end{bmatrix}
\end{align}


 

Without yaw, these dynamics decouple into three 2D subsystems, $(r_x, s_{vx}), (r_y, s_{vy}),$ and $(r_z, s_{vz})$, and we use the technique in \cite{Chen2016DecouplingJournal} to solve \eqref{eq:value_2} using \ref{eq:fastrack_value_1} independently for each 2D subsystem using grid-based techniques. 
Fig.~\ref{fig:6D_decoupled_x} shows the level sets of the value function $V^*$ and corresponding optimal tracker control policies. Fig.~\ref{subfig:grid_x} is the grid-based ground truth, while Fig.~\ref{subfig:nn_optdst_x} shows the induced value function for the neural network classifier policy $\Pi_{-\infty}^u$ trained against the optimal disturbance policy $d^*(\cdot)$, and Fig.~\ref{subfig:nn_lrndst_x} shows the induced value function when $\Pi_{-\infty}^u$ and $\Pi_{-\infty}^d$ were trained jointly. Note that the classification-based results shown here did \emph{not} take advantage of system decoupling. Corroborating our theoretical results, the level sets of the value function induced by our learned classifiers over-approximate the true level sets when the disturbance plays optimally (Fig.~\ref{subfig:fastrack_overlay}). Also, observe that using a learned (and hence, generally suboptimal) $\Pi_{-\infty}^d$, the resulting level sets in Fig.~\ref{subfig:nn_lrndst_x} still well-approximate (though they do not include) those in \ref{subfig:grid_x}.  For each level curve, the maximum tracking error $x$ is the largest value of the level curve along the $r_x$ axis. Observe in Fig.~\ref{subfig:fastrack_overlay} that the maximum tracking error is similar in all three cases. Finally, the line that separates the colored areas in the background of each figure in  Fig.~\ref{fig:6D_decoupled_x} denotes the decision boundary for the controller in each case. 






\subsubsection{7D Coupled}
\label{subsubsec:7D_coupled}

In this example, we introduce yaw ($\yaw$) into the model as an extra state in \eqref{eq:Quad7D_coupled} and introduce yaw rate control $\dot{\yaw} \in [-1.0, 1.0]$ rad/s. The relative position dynamics in $(\rstate_x, \rstate_y, \rstate_z)$ are identical to \eqref{eq:Quad6D_decoupled_relative}. The remaining states evolve as:
\begin{equation}
\small
\label{eq:Quad7D_coupled}
\begin{aligned}
\begin{array}{c}
\left[
\begin{array}{c}
\dot{\tstate}_{vx}\\
\dot{\tstate}_{vy}\\
\dot{\tstate}_{vz}\\
\dot{\tstate}_{\yaw}
\end{array}
\right]
=
\left[
\begin{array}{c}
g (\sin\pitch \cos\tstate_{\yaw} + \sin\roll \sin\tstate_{\yaw})\\
g (-\sin\roll \cos\tstate_{\yaw} + \sin\pitch \sin\tstate_{\yaw})\\
\thrust \cos\roll \cos\pitch - g\\
\dot{\yaw}
\end{array}
\right]
\end{array}
\end{aligned}
\end{equation}


This dynamical model is now \textit{7D}. It is too high-dimensional and coupled in the controls for current grid-based HJ reachability schemes, yet our proposed method is still able to compute a safety controller and the associated largest tracking error. 

\begin{figure}[ht!]
\centering
\includegraphics[width=0.8\columnwidth]{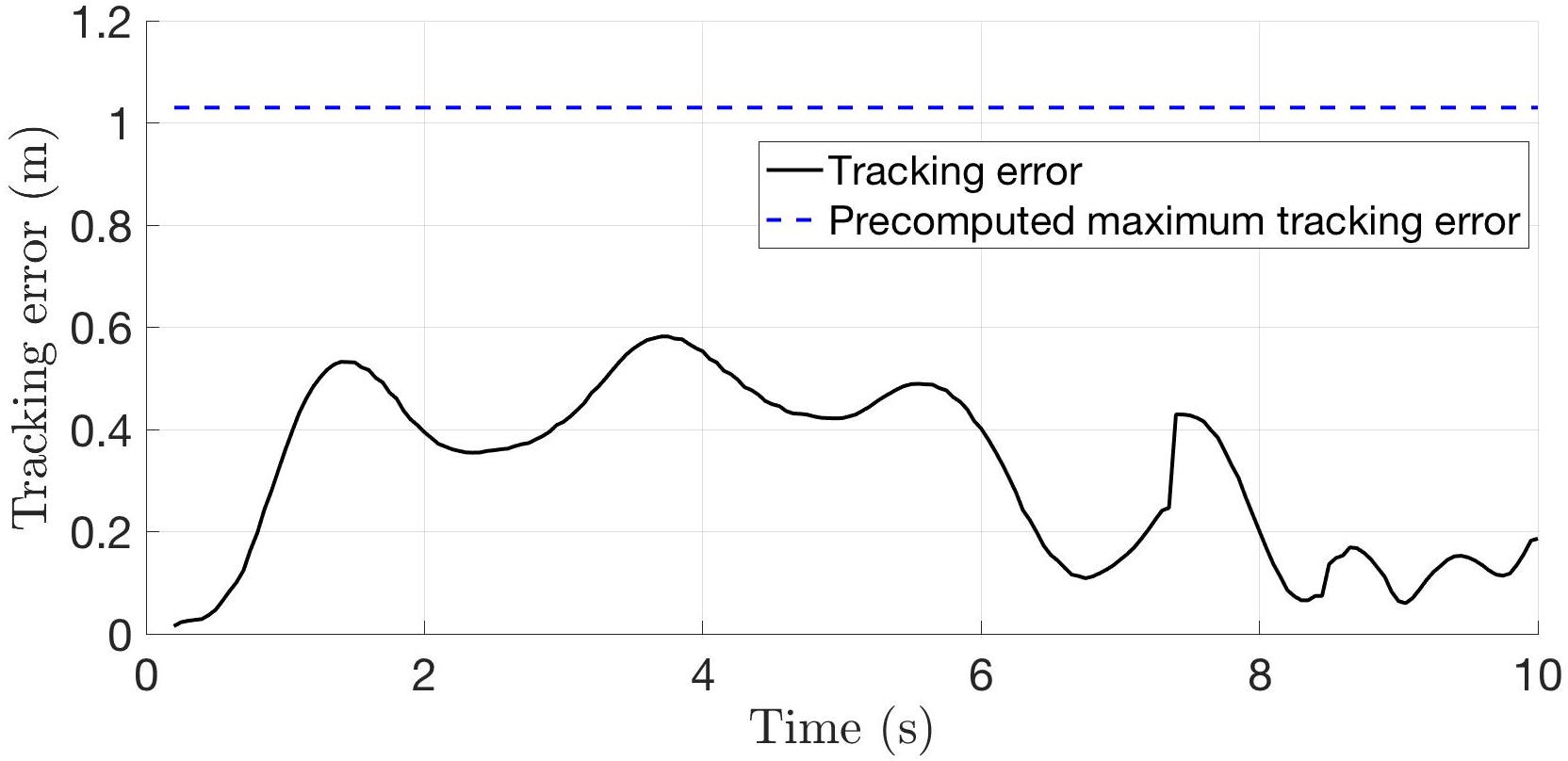}
\caption{Relative distance between the quadrotor (tracking model) and planned trajectory (planning model) over time during a hardware test wherein a Crazyflie 2.0 must navigate through a motion capture arena around spherical obstacles. The quadrotor stays well within the computed tracking error bound throughout the flight. Note that the tracking error is large because our controller accounts for adversarial disturbances, unlike many common controllers.
}
\label{fig:hw_plot}
\vspace{-0.5 cm}
\end{figure}

\subsection{Hardware Demonstration}
\label{subsec:hw_demo}

We tested our learned controller on a Crazyflie 2.0 quadrotor in a motion capture arena.  Fig.~\ref{fig:hw_plot} displays results for \eqref{eq:Quad7D_coupled}. As shown, the quadrotor stays well within the computed error bound.
For this experiment $\Pi_u$ was trained using a sub-optimal disturbance policy.
Even though we do not have a rigorous safety guarantee in this general case because we computed the disturbance, these results corroborate our intuition from Fig.~\ref{fig:6D_decoupled_x} where the computed error bound remains essentially unchanged when using a learned disturbance instead of the optimum. However, by Prop.~\ref{prop:guarantees}, with the optimal disturbance we could compute a strict guarantee. The hardware demonstration can be seen in our video: \url{https://youtu.be/_thXAaEJYGM}. 


\section{Conclusion}
\label{sec:conclusion}

In this paper, we have presented a classification-based approach to approximate the optimal controller in HJ reachability for control-affine systems. We have shown its efficacy in simulation on 2D and 4D environments for reach-avoid problems, and also in a real-time safe trajectory following task involving a 7D quadrotor model. When the optimal disturbance policy is known \textit{a priori}, our method is guaranteed to over-approximate the value function and may thus be used to provide safety and/or goal satisfaction certificates.



\begin{figure}[ht!]
\centering
\includegraphics[width=\columnwidth]{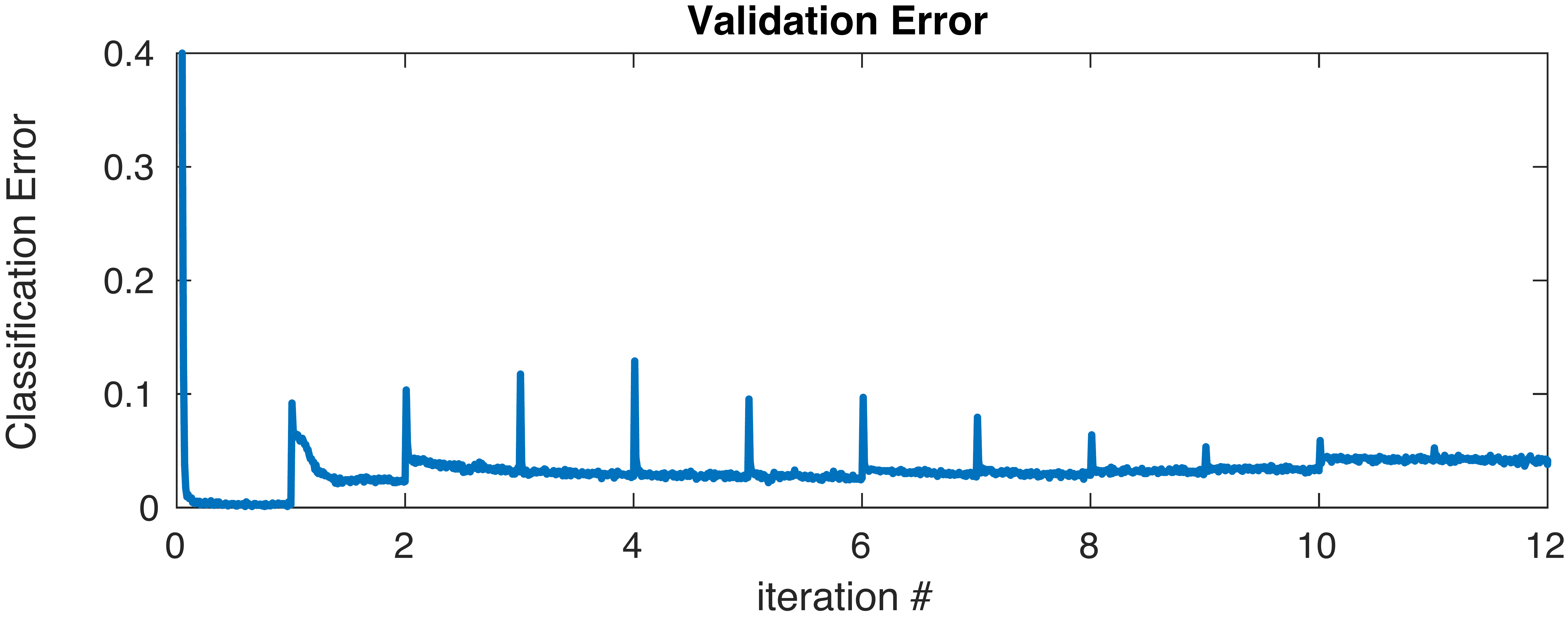}
\caption{Learning curves for a single classifier of the 6D decoupled system. Classification error decreases between spikes, which mark each new $k$ in Algorithm~\ref{alg:PSC}. 
Spikes shrinking hints that classifiers eventually converge.
}
\label{fig:learning_curves}
\vspace{-0.5 cm}
\end{figure}




\section*{APPENDIX}

In this paper we train each binary classifier by minimizing the cross-entropy loss between inputs and labels via stochastic gradient descent. We run the classification problem for a pre-specified number of gradient steps between each new set of policies. Since we expect policies to vary slowly over time, we initialize the weights for each new network with those from its predecessor. This serves two purposes. First, it serves as a ``warm start'' leading to faster stochastic gradient descent convergence. Second, it provides a practical indicator of policy convergence---i.e. if the initial classification accuracy of a new policy is almost equal to that of its predecessor, the policy has most likely converged. Fig. \ref{fig:learning_curves} shows a typical learning curve when running Algorithm \ref{alg:PSC}. The figure shows the progression of the validation error (against unseen state-action pairs) in each iteration. 

All feedforward neural network classifiers had two hidden layers of 20 neurons each, with rectified linear units (ReLUs) as the activation functions, and a final softmax output. The gradient descent algorithm employed was RMSprop with learning rate $\alpha = 0.001$ and momentum constant $\beta = 0.95$. When using function approximators, it is in general unclear how many samples should be taken as a function of the state dimension. In our case, the number of points $N$ sampled at each iteration was $1k$ for the 2D example, and $200k$ for the 4D, 6D and 7D system. All initial weights and biases were drawn from a uniform probability distribution between $[-0.1,0.1]$. All computations were performed on a 12 core, 64-bit machine with Intel® Core™ i7-5820K CPUs @ 3.30GHz. In our implementation we did not employ any form of parallelization. All code for the project can be found at \url{https://github.com/HJReachability/Classification_Based_Reachability}.
 




\balance
\printbibliography
\end{document}